\documentclass[preprint,authoryear,12pt]{elsarticle}

\usepackage{amssymb,epsfig}
\usepackage{psfrag}
\usepackage{amsmath}
\usepackage{adjustbox}
\usepackage{threeparttable,color}

\journal{Medical Image Analysis}

\newcommand{\ct}{{\mathrm {c}}}
\newcommand{\st}{{\mathrm {s}}}

\newcommand{\dd}{{\mathrm d}}

\newcommand{\Cb}{\mathbf{C}}

\newcommand{\Pb}{\mathbf{P}}

\newcommand{\Ib}{\mathbf{I}}

\newcommand{\x}{\mathbf{x}}
\newcommand{\X}{\mathbf{X}}
\newcommand{\y}{\mathbf{y}}

\newcommand{\N}{\mathcal{N}}

\newcommand{\h}{\mathbf{h}}
\newcommand{\W}{\mathbf{W}}
\newcommand{\I}{\mathbf{I}}

\newcommand{\Z}{\mathbf{Z}}

\newcommand{\ab}{\mathbf{a}}
\newcommand{\bb}{\mathbf{b}}
\newcommand{\Rc}{\mathcal{R}}
\newcommand{\F}{\mathbf{F}}
\newcommand{\roi}{\mathrm{roi}}
\newcommand{\shape}{\mathrm{shape}}
\newcommand{\Div}{\mathrm{Div}}

\newcommand{\KL}{\mathrm{KL}}

\newcommand{\p}{\mathbf{p}}
\newcommand{\lv}{\mathrm{lv}}
\newcommand{\lb}{\mathbf{l}}
\newcommand{\addt}[1]{\textcolor{black}{{#1}}}

\begin{document}

\begin{frontmatter}



\title{\addt{A Combined Deep-Learning and Deformable-Model Approach to Fully Automatic Segmentation of the Left Ventricle in Cardiac MRI}}

\author[add1,add2]{M. R. Avendi }
  \ead{m.avendi@uci.edu}
  \author[add2]{Arash Kheradvar}
  \ead{arashkh@uci.edu}
  \author[add1]{Hamid Jafarkhani \corref{cor1} }
    \ead{hamidj@uci.edu}

  \address[add1]{Center for Pervasive Communications and Computing, University of California, Irvine, USA}
  \address[add2]{the Edwards Lifesciences Center for advanced cardiovascular technology,
University of California, Irvine, USA}

\begin{abstract}
Segmentation of the left ventricle (LV) from cardiac magnetic resonance imaging (MRI) datasets is an essential step for calculation of clinical indices such as ventricular volume and ejection fraction. In this work, we employ deep learning algorithms combined with deformable models to develop and evaluate a fully automatic segmentation tool for the LV from short-axis cardiac MRI datasets. The method employs deep learning algorithms to learn the segmentation task from the ground true data. Convolutional networks are employed to automatically detect the LV chamber in MRI dataset. Stacked autoencoders are utilized to infer the shape of the LV. The inferred shape is incorporated into deformable models to improve the accuracy and robustness of the segmentation. We validated our method using 45 cardiac MR datasets taken from the MICCAI 2009 LV segmentation challenge and showed that it outperforms the state-of-the art methods. Excellent agreement with the ground truth was achieved. Validation metrics, percentage of good contours, Dice metric, average perpendicular distance and conformity, were computed as $96.69\%$, 0.94, 1.81mm and 0.86, versus those of $79.2\%-95.62\%$, 0.87-0.9, 1.76-2.97mm and 0.67-0.78, obtained by other methods, respectively. 
\end{abstract}


\end{frontmatter}

\section{Introduction}
\label{sec:intro}
Cardiac magnetic resonance imaging (MRI) is now routinely being used for the evaluation of the function and structure of the cardiovascular system \citep{cmr-yuan-2002,cmr-2004,frangi-review,review-petit,review-tavakoli,review-heiman,collab2014}. Segmentation of the left ventricle (LV) from cardiac MRI datasets is an essential step for calculation of clinical indices such as ventricular volume, ejection fraction, left ventricular mass and wall thickness as well as analyses of the wall motion abnormalities. 

Manual delineation by experts is currently the standard clinical practice for performing the LV segmentation. However, manual segmentation is tedious, time consuming and prone to intra- and inter-observer variability \citep{frangi-review,review-petit,review-tavakoli,review-heiman,collab2014}. To address this, it is necessary to reproducibly automate this task to accelerate and facilitate the process of diagnosis and follow-up. To date, several methods have been proposed for the automatic segmentation of the LV. A review of these methods can be found in \citep{frangi-review,review-petit,review-tavakoli,review-heiman,collab2014}.

To summarize, there are several challenges in the automated segmentation of the LV in cardiac MRI datasets: heterogeneities in the brightness of LV cavity due to blood flow; presence of papillary muscles with signal intensities similar to that of the myocardium; complexity in segmenting the apical and basal slice images; partial volume effects in apical slices due to the limited resolution of cardiac MRI; inherent noise associated with cine cardiac MRI; dynamic motion of the heart and inhomogeneity of intensity; considerable variability in shape and intensity of the heart chambers across patients, notably in pathological cases, etc \citep{review-tavakoli,review-petit,queiros-2014}. Due to these technical barriers the task of automatic segmentation of the heart chambers from cardiac MR images is still a challenging problem \citep{review-petit,review-tavakoli,collab2014}.

Current approaches for automatic segmentation of the heart chambers can be generally classified as: pixel classification \citep{kedenburg2006automatic,cocosco2008automatic}, image-based methods \citep{jolly2009fully,Liu2012}, deformable methods \citep{billet2009cardiac,Ayed-deform-2009,Huadeform2010,stacs-2005}, active appearance and shape models (AAM/ASM) \citep{zhang20104,vanAssen2006286} and atlas models \citep{zhuang2008robust,atlas-lorenzo-2004}. Pixel classification, image-based and deformable methods suffer from a low robustness and accuracy and require extensive user interaction \citep{review-petit}. Alternatively, model-based methods such as AAM/ASM and atlas models can overcome the problems with previous methods and reduce user interaction at the expense of a large training set to build a general model. However, it is very difficult to build a model that is general enough to cover all possible shapes and dynamics of the heart chambers \citep{review-petit,jolly2009}. Small datasets lead to a large bias in the segmentation, which makes these methods inefficient when the heart shape is outside the learning set (e.g., congenital heart defects, post-surgical remodeling, etc).

Furthermore, current learning-based approaches for LV segmentation have certain limitations. For instance, methods using random forests \citep{margeta2012layered,lempitsky2009random,geremia2011spatial} rely on image intensity and define the segmentation problem as a classification task. These methods employ multiple stages of intensity standardization, estimation and normalization, which are computationally expensive and affect the success of further steps. As such, their performance is rather mediocre at basal and apical slices and overall inferior to the state-of-the-art. Also, methods that use Markov random fields (MRFs) \citep{cordero2011unsupervised,huang2004graphical}, conditional random fields (CRFs) \citep{crf_cozas2009,crf_Dreijer2013} and restricted Boltzman machines (RBMs) \citep{fully_carneiro_2014} have been considered. MRF and RBM are generative models that try to learn the probability of input data. Computing the image probability and parameter estimation in generative models is generally difficult and can lead to reduced performance if oversimplified. Besides, they use the Gibbs sampling algorithm for training, which can be slow, become stuck for correlated inputs, and produce different results each time it is run due to its randomized nature. Alternatively, CRF methods try to model the conditional probability of latent variables, instead of the input data. However, they are still computationally difficult, due to complexity of parameter estimation, and their convergence is not guaranteed \citep{crf_Dreijer2013}.

\addt{
Motivated by these limitations, and given the fact that manual segmentation by experts is the ground truth in cardiac MRI, we tackle the complex problem of LV segmentation utilizing a combined deep-learning \citep{lecun2015deep,Hinton-2006,bengio2009learning,bengio2013representation,ufldl-andrew,deng2014deep,baldi2012ae} and deformable-models approach. 
We develop and validate a fully automated, accurate and robust segmentation method for the LV in cardiac MRI. In terms of novelty and contributions, our work is one of the early attempts of employing deep learning algorithms for cardiac MRI segmentation. It is generally believed that since current practices of deep learning have been trained on huge amount of data, deep learning cannot be effectively utilized for medical image segmentation due to the lack of training data. However, we show that even with limited amount of training data, using artificial data enlargement, pre-training and careful design, deep learning algorithms can be successfully trained and employed for cardiac MRI segmentation. Nevertheless, we solve some of the shortcomings of classical deformable models, i.e.,  shrinkage and leakage and sensitivity to initialization, using our integrated approach. Furthermore, we introduce a new curvature estimation method using quadrature polynomials to correct occasional misalignment between image slices. The proposed framework is tested and validated on the MICCAI database \citep{miccai2009}. Finally, we provide better performance in terms of multiple evaluation metrics and clinical indices.
}

The remainder of the manuscript is as follows. In Section \ref{sec:method}, the proposed method is described in detail. In Section \ref{sec:imp_details}, the implementation details are provided. Section \ref{sec:valid_proc} presents the validation experiments. The results are presented in Section \ref{sec:results}. In Section \ref{sec:diss} we discuss the results, performance and comparison with the state-of-the-arts methods. Section \ref{sec:con} concludes the paper.

\section{Materials and Methods}
\label{sec:method}

\subsection{Datasets}
\label{subsec:dataset}
The MICCAI 2009 challenge database \citep{miccai2009} is used in our study to train and assess the performance of the proposed methodology. The MICCAI database was obtained from the Sunnybrook Health Sciences Center, Toronto, Canada. The database is publicly available online \citep{miccai2009} and contains 45 MRI datasets, grouped into three datasets. Each dataset contains 15 cases, divided into four ischemic heart failure cases (SC-HF-I), four non-ischemic heart failure cases (SC-HF-NI), four LV hypertrophy cases (SC-HYP) and three normal (SC-N) cases. Manual segmentation of images by experts at the end diastole (ED) and the end systole (ES) cardiac phases is included in the database. A typical dataset contains 20 frames in 6-12 short-axis slices obtained from the base to the apex. Image parameters are: thickness=8 mm, image size = $256 \times 256$ pixels. 

The training dataset of the MICCAI database \citep{miccai2009} was used to train our method. The validation and online datasets were used for evaluation of the method.

\subsection{Method}
\label{subsec:methods}
\begin{figure}[t!]
\psfrag {Automatic} [c] [] [0.8] {Automatic}
\psfrag {Detection} [c] [] [0.8] {Detection}
\psfrag {step12} [c] [] [1.0] {}

\psfrag {Shape Infer} [c] [] [0.8] {Shape Infer }
\psfrag {Initialization} [c] [] [0.8] {Initialization}
\psfrag {step22} [c] [] [0.7] {\&}

\psfrag {Segmentation} [c] [] [0.8] {\;Segmentation }
\psfrag {step32} [c] [] [0.8] {\&}
\psfrag {Alignment} [c] [] [0.8] {Alignment}

\psfrag {stack of MRI} [c] [] [0.8] {stack of MRI}
\psfrag {LV} [c] [] [0.8] { LV}
\psfrag {shape} [c] [] [1.0] {}

\centerline{\epsfig{figure={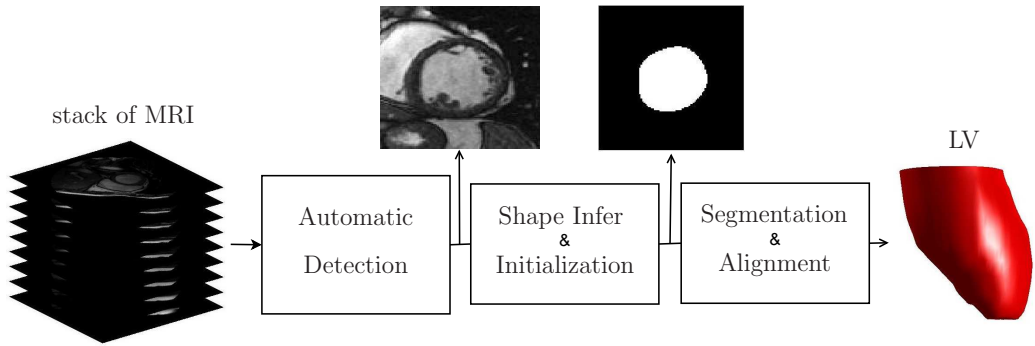},width=\linewidth}}
\caption{{Block diagram of the developed algorithm.}}
\label{fig:bolck}
\end{figure}

The block diagram of the proposed method is depicted in Fig.~\ref{fig:bolck}. A stack of short-axis cardiac MR images is provided as the input (Fig.~\ref{fig:bolck}). The method is carried out in three stages: (i) the region of interest (ROI) containing the LV is determined in the raw input images using convolutional networks \citep{cnn-lecun,cnn_szegedy,cnnlecunn2014,hinton_imagenet_2012} trained to locate the LV; (ii) the shape of the LV is inferred using stacked autoencoders \citep{bengio2013representation,bengio2009learning,ae_vincent,baldi2012ae,deng2014deep,vincent2010stacked} trained to delineate the LV; (iii) the inferred shape is used for initialization and also is incorporated into deformable models for segmentation. Contour alignment is performed to reduce misalignment between slices for 3D reconstruction. Each stage of the block diagram is individually trained during an offline training process to obtain its optimum values of parameters. After training, we deploy the system to perform the automatic segmentation task. 
The three stages are further elaborated as follows:

\subsubsection{Automatic Detection}
\label{sec:autodetect}
The raw cardiac MRI datasets usually include the heart and its surrounding tissues within the thoracic cavity. To reduce the computational complexity and time, and improve the accuracy, the first step of the algorithm is to locate the LV and compute a ROI around it. 

\begin{figure}[t]
\psfrag {m} [c] [] [.6] {64}
\psfrag {m1} [tc] [] [.6] {$\;\;\; 54$}
\psfrag {a} [c] [] [.5] {$11$}
\psfrag {k} [c] [] [.6] {$100$}
\psfrag {m2} [c] [] [.6] {$9$}
\psfrag {Wij} [c] [] [.8] {$\W_1$}
\psfrag {Fl} [c] [] [.7] {$\;\; \F_l$}
\psfrag {m3} [tc] [] [.6] {$32$}
\psfrag {m5} [tc] [] [.6] {$\quad 100$}
\psfrag {p} [c] [] [.6] {$6$}
\psfrag {input} [c] [] [.8] {Input Image}
\psfrag {output} [c] [] [.8] {Output}
\psfrag {ROI} [c] [] [.8] {}
\psfrag {convolutions} [c] [] [.7] {Convolutions}
\psfrag {pooling} [c] [] [.7] {Subsampling}
\psfrag {pooled} [c] [] [.7] {Pooled}
\psfrag {convolved} [c] [] [.7] {Convolved}
\psfrag {features1} [c] [] [.7] {Features}
\psfrag {features2} [c] [] [.7] {Features}
\psfrag {full connection} [c] [] [.7] {Full Connection}
\centerline{\epsfig{figure={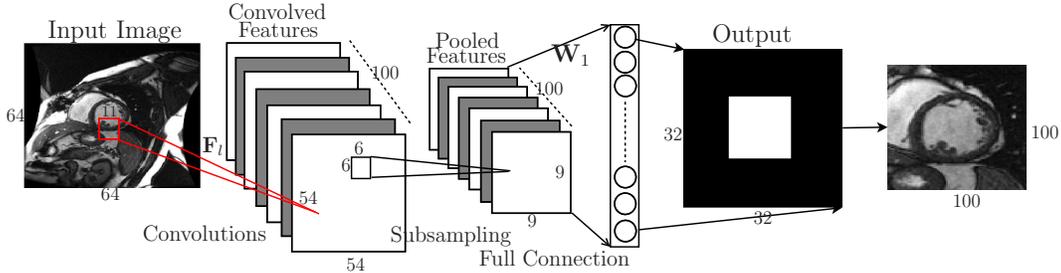},width=\linewidth}}
\caption{{Block diagram of automatic detection of LV in MRI dataset.}}
\label{fig:autodetect}
\end{figure}

A block diagram of the automatic LV detection developed using convolutional networks is illustrated in Fig.~\ref{fig:autodetect}. To reduce complexity, the original image size of $256\times 256$ is down-sampled to $64\times 64$ and used as the input. 

Then, filters $\F_l \in \Rc^{11 \times 11}, \bb_0 \in \Rc^{100}$ are convolved with the input image to obtain the convolved feature maps. Denote the gray-value input image $\Ib:\Omega \rightarrow \Rc, \Omega \subset \Rc^2$ with size $64 \times 64$. $\I[i,j]$ represents a pixel intensity at coordinate $[i,j]$ of the image. Note that the pixel coordinates at the top left and bottom right of the image are $[1,1]$ and $[64,64]$, respectively. The convolved features are computed as $\Cb_l[i,j]= f \left(\Z_l[i,j]\right)$ where
\begin{equation}
\label{eq:conv}
\Z_l[i,j]=\sum \limits_{k_1=1}^{11} \sum \limits_{k_2=1}^{11} \F_l[k_1,k_2] \I[i+k_1-1,j+k_2-1] + \bb_0[l],
\end{equation}
for $1\leq i,j \leq 54$ and $l=1,\cdots,100$. This results in 100 convolved features $\Z_l \in \Rc^{54 \times 54}$.
Here, $\x[i]$ denotes the $i$-th element of vector $\x$ and $\X[i,j]$ denotes the element at the $i$-th row and the $j$-th column of matrix $\X$.

Next, the convolved feature maps are sub-sampled using average pooling \citep{lecun_pool}. To this end, the average values over non-overlapping neighborhoods with size $6\times 6$ are computed in each feature map as
\begin{equation}
\label{eq:Pl}
\Pb_l [i_1,j_1]=\frac{1}{6}  \sum \limits_{i=(6i_1-5)}^{6i_1} \sum \limits_{j=(6j_1-5)}^{6j_1} \Cb_l [i,j],
\end{equation}
for $1 \leq i_1,j_1 \leq 9$. This results in 100 reduced-resolution features $\Pb_l \in \Rc^{9 \times 9}$ for $l = 1,\cdots,100$. 

Finally, the pooled features are unrolled as vector $\p\in \Rc^{8100}$ and fully connected to a logistic regression layer with $1024$ outputs to generate a mask of size $32 \times 32$ that specifies the ROI. The output layer computes $\y_{\ct}=f(\W_1 \p+ \bb_1)$, where $\W_1 \in \Rc^{1024 \times 8100}$ and $\bb_1 \in \Rc^{1024}$ are trainable matrices. 
Note that the original MR image size is $256 \times 256$. Therefore, first, the output mask is up-sampled from $32 \times 32$ to the original MR image size. The center of the mask is then computed and used to crop a ROI of size $100\times 100$ from the original image for further processing in the next stage.

Before using the network for localizing the LV, it should be trained. During training, the optimum parameters of the network ($\F_l,\bb_0,\W_1,\bb_1$) are obtained as described in the next section.

\subsubsection*{\addt{Training Convolutional Network}}
\label{appendix:cnnTrain}
Training the convolution network involves obtaining the optimum values of filters $\F_l, l=1,\cdots,100$ as well as other parameters $\bb_0, \W_1,\bb_1$. In common convolutional networks a large training set is usually available. Therefore, they initialize the filters ($\F_l$) randomly and then train the convolutional network. The filters are constructed simultaneously during training. In our application, the number of training and labeled data is limited. As such, instead of random initialization, the filters are obtained using a sparse autoencoder (AE) which acts as a pre-training step. This leads us to train the network with the limited amount of data that we have while avoid overfitting.  

\begin{figure}[t]
\psfrag {x1} [c] [] [1] {$\x[1]$}
\psfrag {x2} [c] [] [1] {$\x[2]$}
\psfrag {xn} [c] [] [1] {$\x[121]$}

\psfrag {h1} [c] [] [1] {$\ab_2[1]$}
\psfrag {h2} [c] [] [1] {$\ab_2[2]$}
\psfrag {hk} [c] [] [1] {$\ab_2[100]$}

\psfrag {y1} [c] [] [1] {$\y[1]$}
\psfrag {y2} [c] [] [1] {$\y[2]$}
\psfrag {yn} [c] [] [1] {$\y[121]$}
\psfrag {input layer} [c] [] [1] {Input Layer}
\psfrag {output layer} [c] [] [1] {Output Layer}
\psfrag {hidden} [c] [] [1] {Hidden Layer}
\psfrag {w1} [c] [] [1] {$\W_2$}
\psfrag {w2} [c] [] [1] {$\W_3$}
\centerline{\epsfig{figure={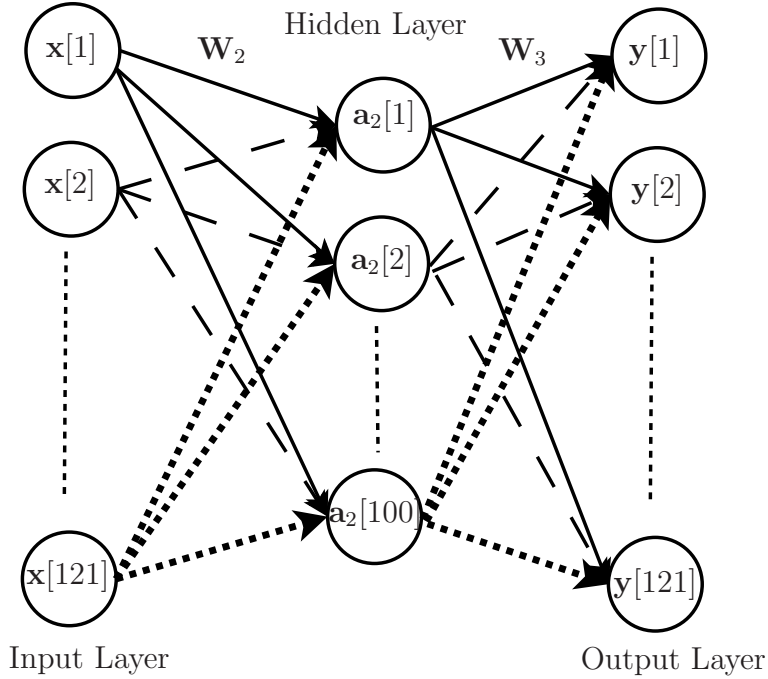},width=.7\linewidth}}
\caption{{Sparse autoencoder is trained to initialize filters ($\F_l$).}}
\label{fig:ae}
\end{figure}

We employ an AE with 121 input/output units and 100 hidden units as depicted in Fig.~\ref{fig:ae}. To train the AE, $N_1 \approx 10^4$ small patches of size $11 \times 11$  are randomly selected from the raw input images of the training dataset. Each patch is then unrolled as vector $\x^{(i)}  \in R^{121}, i=1,\cdots,N_1$ and fed to the input layer of the AE. Denote the weights between the input layer and the hidden layer with $\W_1  \in \Rc^{100 \times 121}$ and the weights between the hidden layer and output layer with $\W_2 \in R^{121 \times 100}$. The hidden layer computes $\ab_2^{(i)}  = f(\W_2 \x^{(i)}+\bb_2)$ and the final output is $\y^{(i)} = f(\W_3 \ab_2^{(i)}+\bb_3)$, where $f(x) = 1/(1+e^{-x})$ is the sigmoid activation function and $\bb_2  \in \Rc^{100}, \bb_3  \in \Rc^{121}$ are trainable bias vectors. The task of AE is to construct $\x^{(i)}$ at the output from the hidden values. Thus, input values are used as the labeled data and no actual labeled data are required for training the AE. 

\begin{figure}[ht!]
\centerline{\epsfig{figure={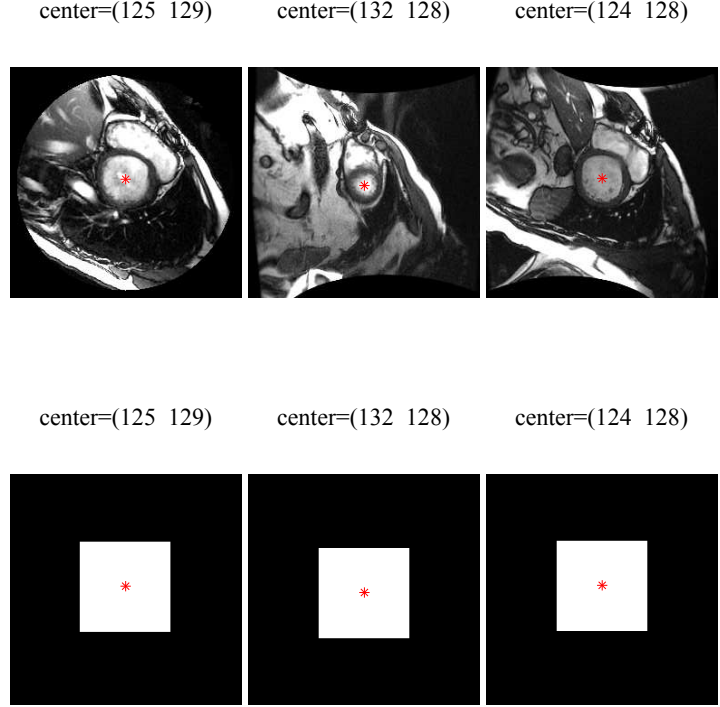},width=.7\linewidth }}
\caption{Typical input images (top) and corresponding binary masks (bottom) used for training of the automatic detection network. Note, the center of image (top) is the same as the center of corresponding ROI (bottom).}
\label{fig:roi_mask}
\end{figure}

The AE is optimized by minimizing the cost function 
\begin{equation}
\label{eq:J_ae}
J(\W_2,\bb_2)=\frac{1}{2N_1} \sum \limits_{i=1}^{N_1}  |\y^{(i)}  - \x^{(i)} |^2 +\frac{\lambda}{2}  \left( \|\W_2\|^2+\|\W_3\|^2 \right)+ \beta \sum \limits_{j=1}^k \KL(\rho || \hat{\rho}_j).
\end{equation}
Here, the first term computes the average squared-error between the final output $\y^{(i)}$ and the desired output $\x^{(i)}$. Furthermore, to avoid overfitting, the $l_2$ regularization/weight decay term is added to the cost function to decrease the magnitude of the weights. Also, to learn higher representation from the input data, a sparsity constraint is imposed on the hidden units. In this way, a sparse AE is built. Here, the Kullback-Leibler (KL) divergence \citep{kullback1951} constrains the mean value of the activations of the hidden layer $\hat{\rho}_j=(1/N_1) \sum \limits_{i=1}^{N_1} \ab_2^{(i)}[j],\; j=1,\cdots,100$, to be equal to the sparsity parameter $\rho$, which is usually a small value. The weight decay coefficient $\lambda$ and the sparsity coefficient $\beta$ control the relative importance of the three terms in the cost function.
The optimization parameters are set as $\lambda=10^{-4},\rho=0.1$ and $\beta=3$. Once autoencodr is trained, $\W_2$ is configured as 100 filters $\F_l \in \Rc^{11 \times 11}, l=1,\cdots,100$ and $\bb_0=\bb_2$ for the next step.

Then, we perform a feed-forward computation using Eqs.~\ref{eq:conv}-\ref{eq:Pl} until the output layer. Next, the output layer is pre-trained by minimizing the cost function
\begin{equation}
\label{eq:J_cnn1}
J(\W_1,\bb_1)=\frac{1}{2N_2} \sum \limits_{i=1}^{N_2} |\y_{\ct}^{(i)}-\lb_{\roi}^{(i)}|^2+ \frac{\lambda}{2}  \left(\| \W_1 \|^2 \right),
\end{equation}
where $\lb_{\roi}^{(i)}\in \Rc^{1024}$ is the labeled data corresponding to the $i$th input image and $N_2$ is the number of training data. The labeled data at the output layer are binary masks, as shown in Fig.~\ref{fig:roi_mask}, generated based on manual training contours. As seen, a binary mask is an image with black background and a white foreground corresponding to the ROI. The foreground is centered at the center of the LV contour, which is known from the training manual contours. Note that the binary mask is down-sampled to $32\times 32$ and then unrolled as vector $\lb_{\roi}^{(i)}$ to be used for training. 

Finally, the whole network is fine-tuned by minimizing the cost function
\begin{equation}
\label{eq:J_cnn}
J(\F_l,\bb_0,\W_1,\bb_1)=\frac{1}{2N_2} \sum \limits_{i=1}^{N_2} |\y_{\ct}^{(i)}-\lb_{\roi}^{(i)}|^2+ \frac{\lambda}{2}  \left(\| \W_1 \|^2+ \sum \limits_{l=1}^{100} \| \F_l \|^2\right)
\end{equation}
The cost function can be minimized using the backpropagation algorithm. Here, $\lambda=10^{-4}$. It should be mentioned that the training process is performed only once.

\begin{figure}
\centerline{\epsfig{figure={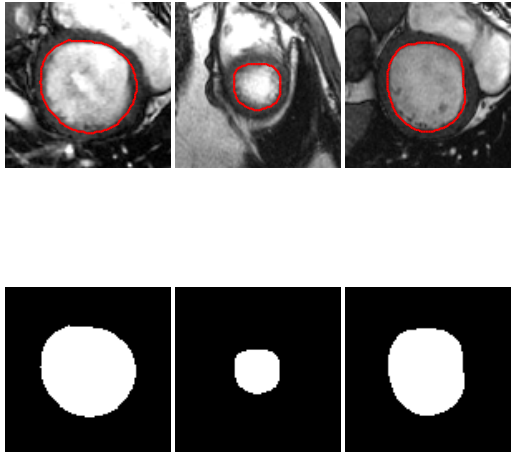}, ,angle=-90,width=.5\linewidth}}
\caption{Typical sub-images with manual segmentation of LV (top) and corresponding binary masks (bottom) used for training the stacked autoencoder.}
\label{fig:lv-masks}
\end{figure}

\subsubsection{Shape Inferring}
\label{sec:shape}
\begin{figure}[t!]
\psfrag {ROI} [c] [] [.8] {ROI}
\psfrag {Inferred Shape} [c] [] [.8] {Inferred Shape}
\psfrag {W4} [c] [] [.8] {$\W_4$}
\psfrag {W5} [c] [] [.8] {$\W_5$}
\psfrag {W6} [c] [] [.8] {$\W_6$}
\psfrag {x1} [c] [] [.8] {$$}
\psfrag {x2} [c] [] [.8] {$$}
\psfrag {xn} [c] [] [.8] {$$}
\psfrag {y1} [c] [] [.8] {$$}
\psfrag {y2} [c] [] [.8] {$$}
\psfrag {yn} [c] [] [.8] {$$}
\psfrag {H1} [c] [] [.8] {$H_1$}
\psfrag {H2} [c] [] [.8] {$H_2$}
\psfrag {Hidden Layers} [c] [] [.8] {Hidden Layers}
\psfrag {Input Layer} [c] [] [.8] {Input Layer}
\psfrag {Output Layer} [c] [] [.8] {Output Layer}
\centerline{\epsfig{figure={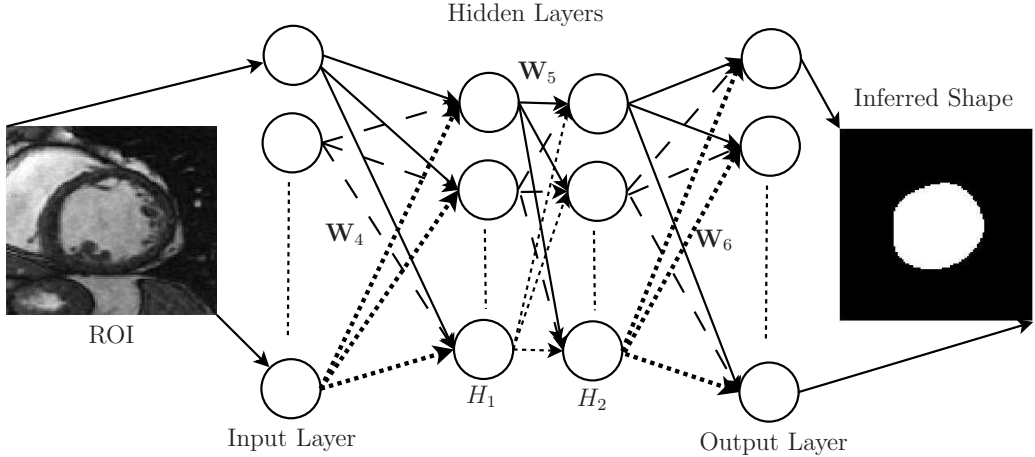}, width=\linewidth}}
\caption{Stacked AE for inferring the shape of LV. The input is a sub-image and the output is a binary mask.}
\label{fig:sae_lv}
\end{figure}

We utilize and train a stacked-AE, depicted in Fig.~\ref{fig:sae_lv}, to infer the shape of the LV. The stacked-AE has one input layer, two hidden layers, and one output layer. 
The sub-image obtained from the previous block is sub-sampled and unrolled as vector $\x_{\st} \in \Rc^{4096}$ and fed to the input layer. The hidden layers build the abstract representations by computing $\h_1=f(\W_4 \x_{\st}+\bb_4)$ and $\h_2=f(\W_5 \h_1+\bb_5)$. The output layer computes $\y_{\st}=f(\W_6 \h_2+\bb_6)$ to produce a binary mask. The binary mask is black (zero) everywhere except at the borders of the LV. Here, $\W_4 \in \Rc^{100 \times 4096}, \bb_4 \in \Rc^{100}$,$\W_5 \in \Rc^{100 \times 100}, \bb_5 \in \Rc^{100}$ and
$\W_6 \in \Rc^{4096 \times 100}, \bb_6 \in \Rc^{4096}$ are trainable matrices and vectors that are obtained during the training process, as detailed in the next section.

\subsubsection*{\addt{Training stacked-AE}}
\label{apped:train-stack-ae}
The training of the stacked-AE is performed in two steps: pre-training and fine-tuning. Since the amount of labeled data is limited in our application, a layer-wise pre-training is performed. The layer-wise pre-training helps to prevent overfitting, leading to a better generalization.  During the pre-training step, parameters $\W_4, \W_5$ of the stacked-AE  are obtained layer by layer with no labeled data. Parameter $\W_6$ of the stacked-AE is obtained using the labeled data. The details are as follows.

First, the input layer and the hidden layer $H_1$ are departed from the stacked-AE. By adding an output layer with the same size as the input layer to the two departed layers (input layer and $H_1$) a sparse AE is constructed (similar to Fig.~\ref{fig:ae}). The sparse AE is trained in an unsupervised fashion as explained in \ref{appendix:cnnTrain} to obtain $\W_4$. The optimization parameters are set as $\lambda=3 \times 10^{-3}, \rho=0.1, \beta=3$. 

The training input/output data of the sparse AE are sub-images of size $100\times 100$  centered at the LV extracted from the full-size training images. The input image is resized to $64\times 64$ to be compatible with the input size $4096$ of the stacked-AE. Once training of the first sparse AE is complete, its output layer is discarded. The hidden units' outputs in the AE are now used as the input for the next hidden layer $H_2$ in Fig.~\ref{fig:sae_lv}. 

Then, hidden layers $H_1$ and $H_2$ are departed from the stacked-AE to construct another sparse AE. Similarly, the second sparse AE is trained to obtain $\W_5$. Again, no labeled data is required. This step can be repeated depending on the number of hidden layers. 

The last hidden layer's outputs are used as the input to the final layer, which is trained in a supervised fashion to obtain $\W_6$. The cost function to train the final layer computes 
\begin{equation}
\label{eq:J_sae_out}
J(\W_6,\bb_6)=\frac{1}{2N_2} \sum \limits_{i=1}^{N_2} |\y_{\st}^{(i)}-\lb_{\lv}^{(i)}|^2+ \frac{\lambda}{2}  \| \W_6 \|^2,
\end{equation}
where $\lb_{\lv}^{(i)} \in \Rc^{4096}$ is the labeled data corresponding to the $i$th image. The labeled data are binary masks created from manual segmentations drawn by experts. Fig.~\ref{fig:lv-masks} depicts three examples of input images and corresponding labels used for training of the stacked-AE. Note that the binary mask is unrolled as vector $\lb_{\lv}$ to be used during optimization.  

It should be mentioned that the layer-wise pre-training results in proper initial values for parameters $\W_4, \W_5, \W_6$. In the second step, the whole architecture is fine-tuned by minimizing the cost function
\begin{multline}
\label{eq:J_sae_finetune}
J(\W_4,\W_5,\W_6,\bb_4,\bb_5,\bb_6) =\\ \frac{1}{2N_2} \sum \limits_{i=1}^{N_2} |\y_{\st}^{(i)}-\lb_{\lv}^{(i)}|^2+ \frac{\lambda}{2} \left( \|  \W_4 \|^2+ \| \W_5 \|^2+\| \W_6 \|^2\right ),
\end{multline}
using the back-propagation algorithm with respect to the supervised criterion. Here $\lambda=10^{-4}$. As in the case of automatic detection the training process is performed only once.

\subsubsection{Segmentation and Alignment}
\label{subsec:seg}
The final block employs a deformable model combined with the inferred shape for accurate segmentation. Deformable models are dynamic contours that evolve by minimizing an energy function. The energy function reaches its minimum when the contour lies on the boundary of the object of interest. In most of conventional deformable methods, the output contours tend to shrink inward or leak outward due to presence of the papillary muscles in the LV and small contrast between surrounding tissues. We solve these issues by using the inferred shape from the previous stage as a good initialization. In addition, the shape is incorporated into the energy function to prevent the contour from shrinkage/leakage. 

Denote the input sub-image with $I_{\st}: \Omega_{\st} \rightarrow \Rc, \Omega_{\st} \subset \Omega \subset \Rc^2$ and the coordinate of image pixels with $(x,y)$. Let us define $\phi(x,y)$ as the level set function that returns negative values for the pixels inside a contour and positive values for the pixels outside. Also, denote the level set function corresponding to the inferred shape with $\phi_{\shape}(x,y)$. The energy function is defined as 
\begin{equation}
\label{eq:E}
E(\phi)= \alpha_1 E_{\mbox{len}}(\phi)+ \alpha_2 E_{\mbox{reg}}(\phi)+\alpha_3 E_{\mbox{shape}}(\phi),
\end{equation}
which is a combination of the length-based energy function \citep{ac_wo_edges}
\begin{equation}
\label{eq:E_leng}
E_{\mbox{len}}(\phi)= \int_{\Omega_{\st}}  \delta(\phi) |\nabla \phi| \dd x \dd y, 
\end{equation}
region-based \citep{ac_wo_edges}
\begin{equation}
\label{eq:Ereg}
E_{\mbox{reg}}(\phi)=\int_{\Omega_{\st}} |I_{\st}-c_1|^2 H(\phi) \dd x \dd y + \int_{\Omega_{\st}} |I_{\st}-c_2|^2 (1-H(\phi)) \dd x \dd y,
\end{equation}
and prior shape energy terms
\begin{equation}
\label{eq:Eshape}
E_{\mbox{shape}}(\phi)= \int_{\Omega_{\st}} (\phi - \phi_{\mbox{shape}})^2 \dd x \dd y.
\end{equation}
Here, $\delta(\phi)$, $H(\phi)$ and $\nabla(\cdot)$ are the delta function, Heaviside step function and the gradient operation, respectively. Also $c_1$ and $c_2$ are the average of the input image $I_{\st}$ outside and inside the contour \citep{ac_wo_edges}, respectively. \addt{The $\alpha_i$'s, $i = 1,\cdots, 3 $ are the combining parameters, which were determined empirically during training as $\alpha_1=1, \alpha_2=0.5, \alpha_3=0.25$.}

The deformable method seeks a unique contour denoted by $C^*$ (or equivalently $\phi^*$), which lies on the boundary of the object of interest. This is obtained by minimizing the energy function over $\phi$ as:
\begin{equation}
\label{eq:phi*}
\phi^*= \arg \min \limits_{\phi} \{ E(\phi)\},
\end{equation}
that can be solved using the gradient descent algorithm. By letting $\phi$ be a function of time and using the Euler-Lagrange equation \citep{ac_wo_edges,stacs-2005}, we obtain
\begin{multline}
\label{eq:PDE}
\frac{\dd \phi}{\dd t}= - \frac{\dd E}{\dd \phi}=\delta(\phi) 
 \left[ \alpha_1 \Div\left(\frac{\nabla \phi}{|\nabla \phi|} \right) + \alpha_2 (I_{\st}-c_2)^2 \right. \\ \left. -  
 \alpha_2 (I_{\st}-c_1)^2-2 \alpha_3 (\phi - \phi_{\shape}) \right],
\end{multline}
where $\Div(\cdot)$ is the divergence operator. 

The gradient descent starts with an initialization of $\phi^{(0)}$ obtained from the inferred shapes and is updated iteratively 
\begin{equation}
\label{eq:phikp1}
\phi^{(k+1)}=\phi^{(k)}+ \gamma \frac{\dd \phi}{\dd t},
\end{equation}
to obtain the final $\phi^*$ or contour $C^*$. Here, $\gamma$ is the step size which is a small number. The stopping criterion checks whether the solution is stationary or not by computing the difference between the length of the contours in the current and previous iterations.

\addt{
In case of 3D reconstruction of cardiac chambers, it is necessary to consider possible misalignment between the image slices. Misalignment occurs in cardiac MRI mainly due to respiratory and patient motions during MRI scans. Ignoring misalignment leads to jagged discontinuous surfaces in the reconstructed volume. To deal with this issue, we introduce a misalignment estimation and correction using quadratic polynomials.} 

\addt{
To this end, the center coordinate of the LV contours is computed from the obtained LV segmentation in all image slices, denoted as $(\tilde{x}_i,\tilde{y}_i)$, for $i=1,\cdots,n$, where $n$ is the number of slices. Let us denote the actual center coordinate of the $i$th contour with $(x_i,y_i)$. Then we can write
\begin{gather}
\label{eq:align}
\tilde{x}_i=x_i+w_i,\\
\tilde{y}_i=y_i+v_i,
\end{gather}
where $w_i \sim \N(0,\sigma_w^2)$, $v_i \sim \N(0,\sigma_v^2)$ are the misalignment values due to motion artifacts, modeled by independent Gaussian random variables.}

\begin{figure}[h!]
\psfrag {X} [c] [] [1.0] {$X$}
\psfrag {Y} [c] [] [1.0] {$Y$}
\psfrag {Z} [c] [] [1.0] {$Z$}
\psfrag {misaligned} [c] [] [0.8] {\qquad Misaligned}
\psfrag {aligned} [c] [] [0.8] {\qquad Aligned}
\centerline{\epsfig{figure={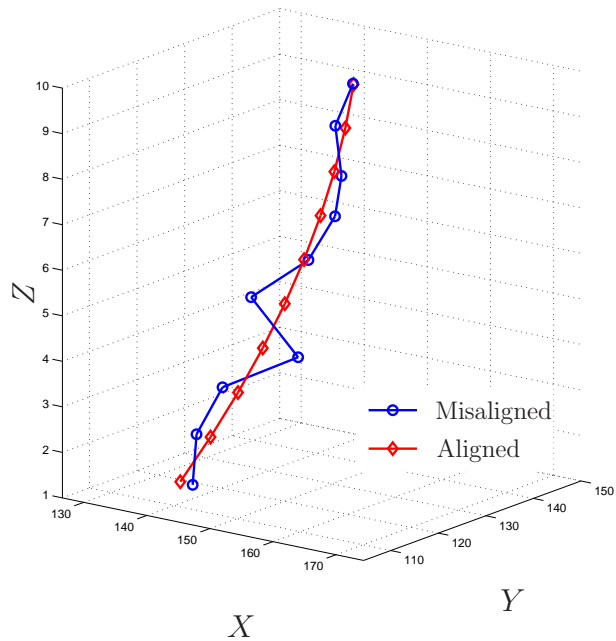},width=\linewidth}}
\caption{Misaligned centers of LV contours from the base to the apex (blue) and corresponding aligned centers (red) obtained from quadrature curve fitting for a typical MRI dataset with ten image slices.}
\label{fig:aligned-misaligned}
\end{figure}

\addt{
Using quadratic assumption for the curvature, it follows that
\begin{gather}
\label{eq:xi}
x_i= a_1 i^2 + b_1 i+ c_1, \\
\label{eq:yi}
y_i= a_2 i^2 + b_2 i+ c_2. 
\end{gather}
Here $a_1,b_1,c_1,a_2,b_2,c_2$ are unknown parameters that are estimated by minimizing the mean squared error as
\begin{gather}
\hat{a}_1,\hat{b}_1,\hat{c}_1= \arg \min_{a_1,b_1,c_1} \sum \limits_{i=1}^n (\tilde{x_i}-a_1 i^2 - b_1 i- c_1)^2, \\
\hat{a}_2,\hat{b}_2,\hat{c}_2= \arg \min_{a_2,b_2,c_2} \sum \limits_{i=1}^n (\tilde{y_i}-a_2 i^2 - b_2 i- c_2)^2.
\end{gather}
After estimating the unknown parameters, the actual center coordinates are estimated from Eqs.~(\ref{eq:xi})-(\ref{eq:yi}).
Finally, the contours are registered, using an affine transformation with linear interpolation, according to the estimated center values to obtain an aligned stack of contours.
Fig.~\ref{fig:aligned-misaligned} illustrates the centers of LV contours from the base to the apex for a typical MRI dataset with ten misaligned image slices. The estimated aligned centers using quadratic polynomials are depicted in red in the figure.}

\section{Implementation Details}
\label{sec:imp_details}
\addt{
Images and contours of all the cases in the training dataset of the MICCAI challenge \citep{miccai2009} were collected and divided into the large-contour and small-contour groups. Typically, the large contours belong to image slices near the base/middle and the small contours belong to the apex of the heart since the contours near the apex of the heart are much smaller than the contours at the base. As such, there are around 135 and 125 images in each group, respectively. Then, we artificially enlarged the training dataset using techniques such as image translation, rotation and changing the pixel intensities based on the standard principal component analysis (PCA) technique as explained in \citep{dataenlarge}. Using these techniques, we augmented the training dataset by a factor of ten. Eventually, we had around 1350 and 1250 images/labels in each group, respectively. Then, we built and trained two networks, one for the large-contour dataset and one for the small-contour dataset.}

It is noted that considerable overfitting may happen in deep learning networks, due to the large number of parameters to be learned. We paid great attention to prevent the overfitting problem in our networks. To deal with this, we adopted multiple techniques including: layer-wise pre-training, $l_2$ regularization and sparsity constraints as explained in Sections \ref{sec:autodetect} and \ref{sec:shape}. Although the lack of training data was a challenge, the use of layer-wise pre-training was greatly helpful. We also kept the number of units in the hidden layers small and did not go beyond three layers to ensure that the number of parameters is tractable. Furthermore, we performed cross-validation and early stopping to monitor and prevent overfitting. In addition, we artificially enlarged the training dataset as mentioned earlier in this section. The hyper-parameters of the networks, i.e., number of layers and units, number of filters,  filter and pooling sizes, etc., are determined empirically during the cross-validation process.  

In the current study, our method was developed in MATLAB 2014a, performed on a Dell Precision T7610 workstation, with Intel(R) Xeon(R) CPU 2.6 GHz, 32 GB RAM, 64-bit Windows 7. The method was trained using the training dataset and tested on the online and validation datasets of the MICCAI database \citep{miccai2009}.

\begin{figure}[t!]
\begin{minipage}{.5\linewidth}
\centerline{\epsfig{figure={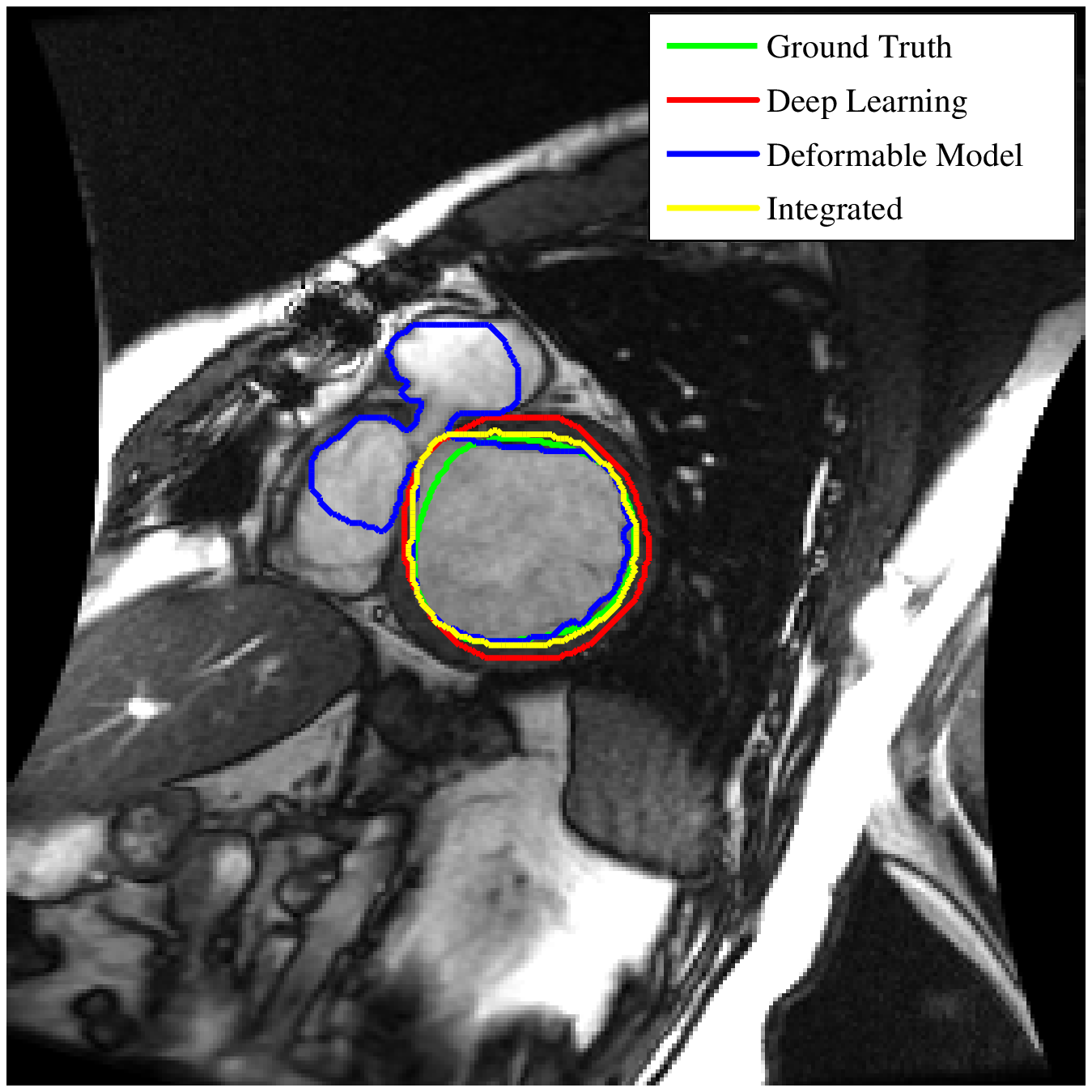},width=\linewidth}}
\end{minipage}
\begin{minipage}{.5\linewidth}
\centerline{\epsfig{figure={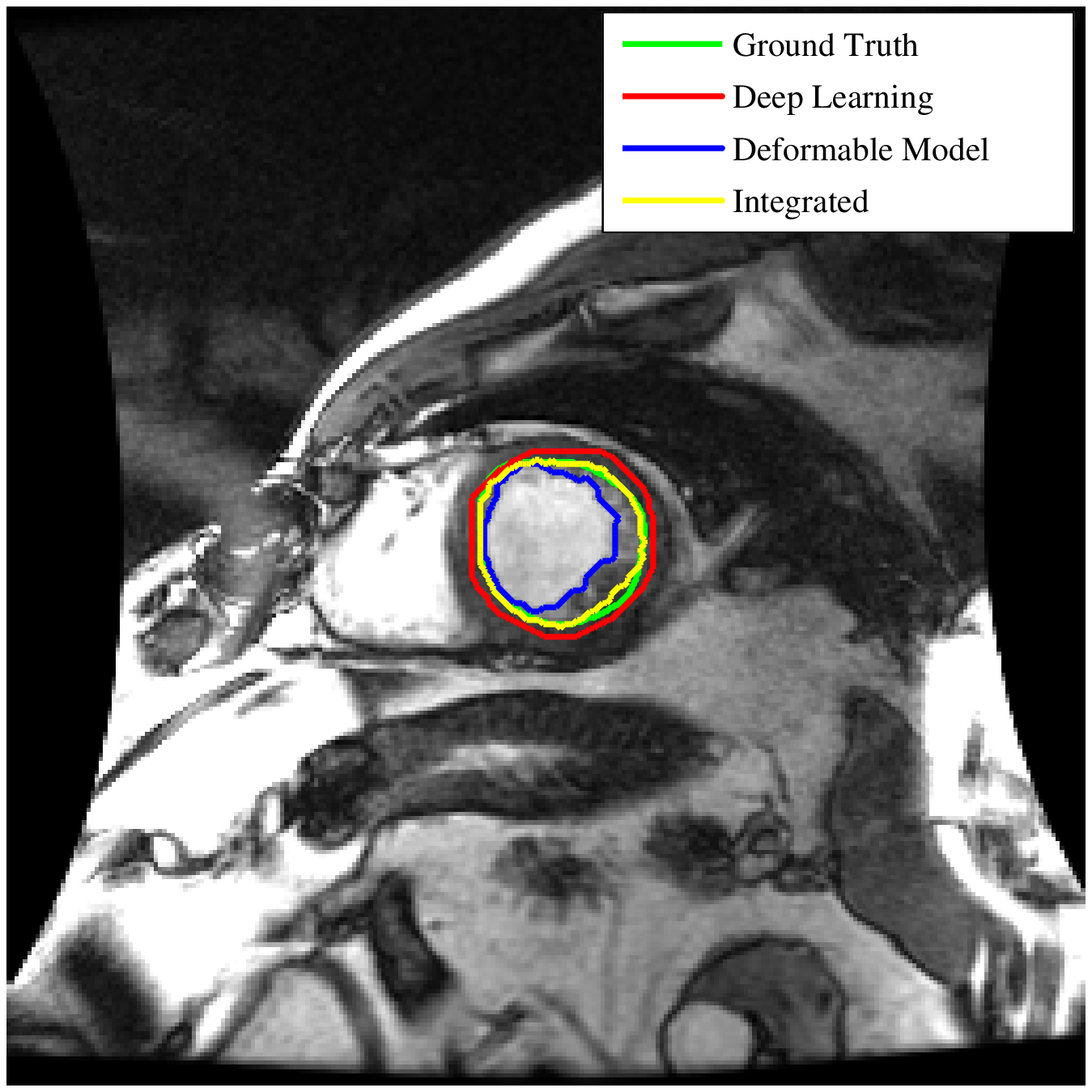},width=\linewidth}}
\end{minipage}
\caption{\addt{Outcomes of deformable model with no shape constraint ($\alpha_3=0$), deep learning (shape inference, step 2) and integrated deep learning and deformable model (final step), for two typical images.}}
\label{fig:each_step}
\end{figure}

\begin{figure}[t!]
\psfrag {input} [c] [] [1.0] {Input Image}
\psfrag {stackedae} [c] [] [1.0] {Deep Architecture}
\psfrag {output} [c] [] [1.0] {Inferred Shape}
\centerline{\epsfig{figure={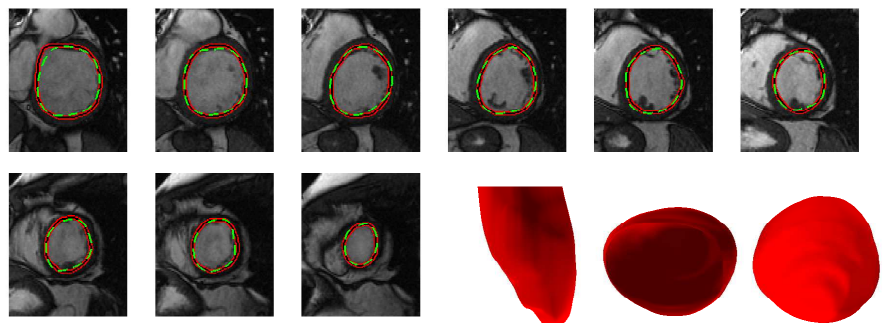},width=\linewidth}}
\caption{Automatic (red-black) and manual (dashed green) segmentation results of LV for an example cardiac MRI dataset of the MICCAI database \citep{miccai2009} in 2D and 3D (right) representations.}
\label{fig:segs2d3d}
\end{figure}

\begin{figure}[h]
\psfrag {input} [c] [] [1.0] {Input Image}
\psfrag {stackedae} [c] [] [1.0] {Deep Architecture}
\psfrag {output} [c] [] [1.0] {Inferred Shape}
\centerline{\epsfig{figure={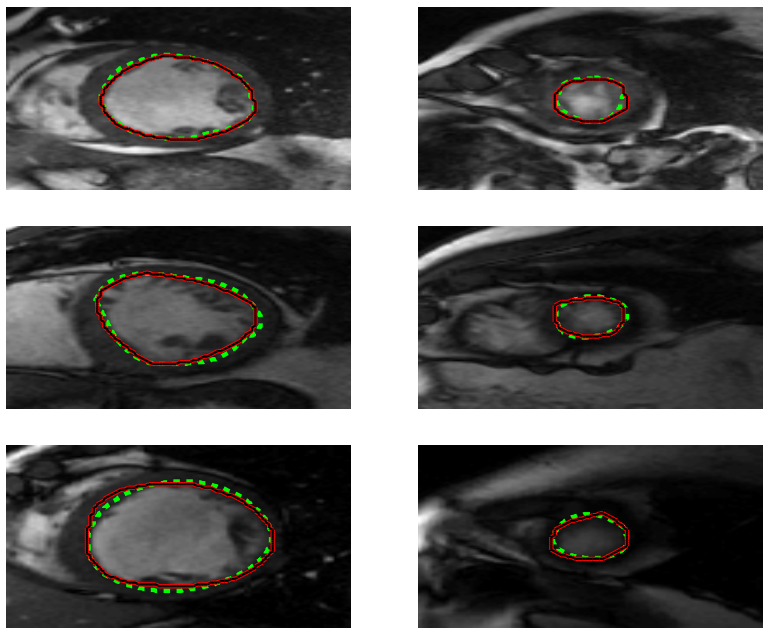},width=.8\linewidth}}
\caption{Automatic (red-black) and manual (green) segmentation results for challenging images at the apex (right) and mid LV (left) with presence of papillary muscles for three typical cardiac MRI datasets of the MICCAI database \citep{miccai2009}.}
\label{fig:challenging}
\end{figure}

\section{Validation Process}
\label{sec:valid_proc}
We assess the performance of the proposed methodology by evaluating the accuracy of the proposed automated segmentation method compared with the gold standard (manual annotations by experts). To this end, the following measures are computed as: average perpendicular distance (APD), Dice metric, Hausdorff distance, percentage of good contours and the conformity coefficient \citep{chang2009performance}. As recommended in \citep{miccai2009}, a segmentation is classified as good if the APD is less than 5mm. The average perpendicular distance measures the distance from the automatically segmented contour to the corresponding manually drawn expert contour, averaged over all contour points \citep{miccai2009}. A high value implies that the two contours do not match closely \citep{miccai2009}.  Also, the Dice metric, $\mbox{DM}=2(A_{\mbox{am}})/(A_{\mbox{a}}+A_{\mbox{m}})$, is a measure of contour overlap utilizing the contour areas automatically segmented $A_{\mbox{a}}$, manually segmented $A_{\mbox{m}}$, and their intersection $A_{\mbox{am}}$ \citep{miccai2009}. The Dice metric is always between zero and one, with higher DM indicating better match between automated and manual segmentations. The Hausdorff distance measures the maximum perpendicular distance between the automatic and manual contours \citep{queiros-2014,eval_segs}. Finally, the conformity coefficient measures the ratio of the number of mis-segmented pixels to the number of correctly segmented pixels defined as CC=(3DM-2)/DM \citep{chang2009performance}. 

In addition, three clinical parameters, end-diastolic volume (EDV), end-systolic volume (ESV) and ejection fraction (EF) were computed using the automatic and manual LV segmentation results and used for the correlation and Bland-Altman analyses \citep{bland1986statistical}. The correlation analysis was performed using the Pearson’s test to obtain the slope and intercept equation and the Pearson $R$-values. To assess the intra- and inter-observer variability the coefficient of variation (CV), defined as the standard deviation (SD) of the differences between the automatic and manual results divided by their mean values, and the reproducibility coefficient (RPC), defined as $1.96*$SD, are computed. 

The segmentation performance was assessed against reference contours using the evaluation code provided in \citep{miccai2009}. Each measure is computed slice by slice and a mean value and standard deviation for all slices of a dataset are calculated. 

\section{Results}
\label{sec:results}

\begin{figure}[t!]
\psfrag {base} [c] [] [1.0] {Base}
\psfrag {apex} [c] [] [1.0] {Apex}

\psfrag {p1} [c] [] [1.0] {SC-HF-I\;\;}
\psfrag {p2} [c] [] [1.0] {SC-HF-NI \quad}
\psfrag {p3} [c] [] [1.0] {SC-HYP\quad}
\psfrag {p4} [c] [] [1.0] {SC-N}

\psfrag {p5} [c] [] [1.0] {SC-HF-I\quad}
\psfrag {p6} [c] [] [1.0] {SC-HF-NI\qquad}
\psfrag {p7} [c] [] [1.0] {SC-HYP\qquad}
\psfrag {p8} [c] [] [1.0] {SC-N \quad}

\psfrag {p9} [c] [] [1.0] {SC-HF-I\quad}
\psfrag {p10} [c] [] [1.0] {SC-HF-NI\qquad}
\psfrag {p11} [c] [] [1.0] {SC-HYP\qquad}
\psfrag {p12} [c] [] [1.0] {SC-N\quad}

\psfrag {dt1} [c] [] [1.0][-90] {Validation}
\psfrag {dt2} [c] [] [1.0] [-90] {Online}
\psfrag {dt3} [c] [] [1.0] [-90] {Training}

\centerline{\epsfig{figure={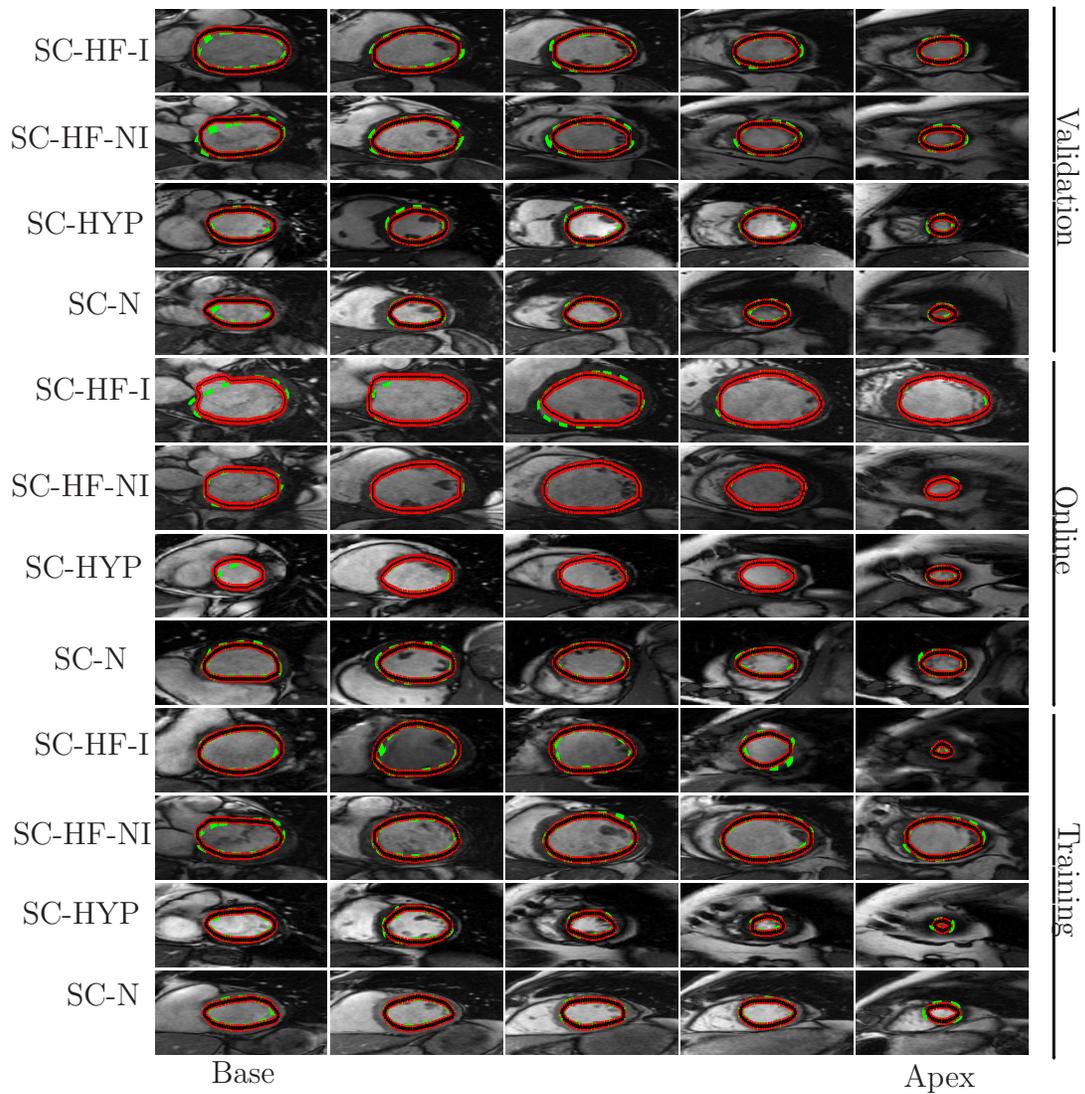},height=1.2\linewidth,width=\linewidth}}
\vspace{-.65in}
\caption{Automatic (red-black) and manual (dashed green) segmentation of LV in the base (left), mid-ventricular (middle) and the apex (right) slices for multiple cases of the MICCAI database \citep{miccai2009}. Each row corresponds to one patient, ischemic heart failure (SC-HF-I), non-ischemic heart failure (SC-HF-NI), LV hypertrphies (SC-HYP) and normal (SC-N)}
\label{fig:seg_multiplecases}
\end{figure}

\subsection{Illustrative Results}
\label{sec:illresults}
\addt{To better understand the role of each step, the outcome of the deformable model with no shape constraint ($\alpha_3=0$), deep learning (shape inference, Step 2) and the integrated deformable model and deep learning method (final step) for two typical images are shown in Fig.~\ref{fig:each_step}.}

Fig.~\ref{fig:segs2d3d} illustrates the automatic and manual segmentation results of the LV for a typical cardiac MRI dataset from the base to the apex and three views of the reconstructed LV chamber (front, base and apex views). Also, segmentation results for image slices at the apex and mid LV, which are generally complex due to presence of papillary muscles and low resolution, are depicted in Fig.~\ref{fig:challenging}. In the figures, automatic segmentation results are shown in red. The ground truth manual segmentations drawn by experts are shown in green for comparison. Automatic and manual segmentation results for multiple datasets of the MICCAI database \citep{miccai2009} are illustrated in Fig.~\ref{fig:seg_multiplecases}. In the figure, each row corresponds to one patient/dataset which includes normal subjects (SC-N) and the ones with ischemic heart failure (SC-HF-I), non-ischemic heart failure (SC-HF-NI) and LV hypertrophy (SC-HYP).  

\subsection{Quantitative Results}
\label{sec:qresults}

\begin{table}[t]
\caption{Evaluation metrics of our proposed method for the validation and online datasets of the MICCAI database \citep{miccai2009}. Numbers' format: mean value (standard deviation).}
\begin{center}
\begin{adjustbox}{max width=\textwidth}
\begin{threeparttable}
  \begin{tabular}{  l l  l l  l l l  } \hline
{Dataset}  & I/F\tnote{3}  & {{Good Contours (\%)}} &  {{Dice Metric}} & {{APD\tnote{1} (mm)}} & {{HD\tnote{2} (mm)}} & Conformity \\ \hline 
    {{Validation}} & I &  $90 (10)$ & $0.90 (0.1) $ & $2.84 (0.29)$  & $3.29 (0.59)$ & $0.78(0.03)$ \\ 
    {{Validation}} & F &  $97.8 (4.7)$ & $0.94 (0.02) $ & $1.7 (0.37)$  & $3.29 (0.59)$ & $086(0.04)$ \\ 
\\
    {{Online}} & I & $87(12)$ & $0.89(0.03)$ & $2.95(0.54)$  & $4.64(0.76)$ & $0.76(0.07)$ \\ 
    {{Online}} & F & $ 95.58 (6.7)$ & $0.93 (0.02)$ & $1.92 (0.51)$  & $3.62 (1.1)$ & $0.85 (0.05)$ \\ 
    \hline
  \end{tabular}
\begin{tablenotes}
\item[1] Average Perpendicular Distance (APD).
\item[2] Hausdorff Distance (HD).
\item[3] (I): Initial contour, (F) Final contour.
\end{tablenotes}  
\end{threeparttable}
\end{adjustbox}
\end{center}
\label{tab:comp1}
\end{table}

\begin{table}[t]
\caption{ Comparison of segmentation performance between proposed method and state-of-the-art techniques using the MICCAI database \citep{miccai2009}. Numbers’ format: mean value (standard deviation).}
\begin{center}
\begin{adjustbox}{max width=\textwidth}
\begin{threeparttable}
  \begin{tabular}{ l  l  l l l l} \hline
{Method}    & \#\tnote{1} & {{Good Contours(\%)}} &  {{Dice Metric}} & {{APD\tnote{2} (mm)}}  & Conformity \\ \hline 
    \textbf{Proposed} & 30 & $\bf{96.69 (5.7)}$ & $\bf{ 0.94 (0.02)}$  & $\bf{1.81 (0.44)}$ & $\bf{0.86}$    \\ 
    
    {\citep{queiros-2014}} & 45 &  $ 92.7 (9.5)$ & $0.9 (0.05)$ &  $1.76 (0.45)$   & $0.78$ \\ 
    
    {\citep{fully_carneiro_2014}} & 30 &  $ 93.23 (9.84)$ & $0.89 (0.03)$ &  $2.26 (0.46)$  & $0.75$  \\

    { \citep{Hu2013}} & 45 &  $ 91.06 (9.4) $ & $0.89 (0.03)$ &  $2.24 (0.4)$ & $0.75$  \\ 

    { \citep{constant2012}} & 45 &  $ 80 (16) $ & $0.86 (0.05)$ &  $2.44 (0.56)$ & $0.67$ \\ 

    { \citep{Liu2012}} & 45 &  $ 91.17 (8.5) $ & $0.88 (0.03)$ &  $2.36 (0.39)$  & $0.73$  \\ 

    { \citep{Huang2011}} & 45 &  $  79.2 (19)  $ & $0.89 (0.04) $ &  $2.16 (0.46) $  & $0.75$  \\ 
    
       { \citep{Schaerer2010}} & 45 &  $ - $ & $0.87 (0.04) $ &  $2.97 (0.38) $  & $0.70$  \\ 
    
       { \citep{jolly2009fully}} & 30 &  $ 95.62 (8.8) $ & $0.88 (0.04) $ &  $2.26 (0.59) $  & $0.73$  \\ \hline        
  \end{tabular}
\begin{tablenotes}
\item[1] Number of datasets evaluated. 30 -validation and online datasets, 45- all datasets.
\item[2] Average Perpendicular Distance (APD).
\end{tablenotes}  
\end{threeparttable}  
\end{adjustbox}
\end{center}
\label{tab:comp2}
\end{table}

In Table~\ref{tab:comp1}, the average values and the standard deviation of the computed metrics are listed for the validation and online datasets. For each dataset, two rows of results, corresponding to the initial contour (I) obtained from the inferred shape and the final contour (F), are listed. Table~\ref{tab:comp2} presents a comparison of results between our method with the state-of-the-art methods that used the same database.

\begin{figure}[t!]
\centerline{\epsfig{figure={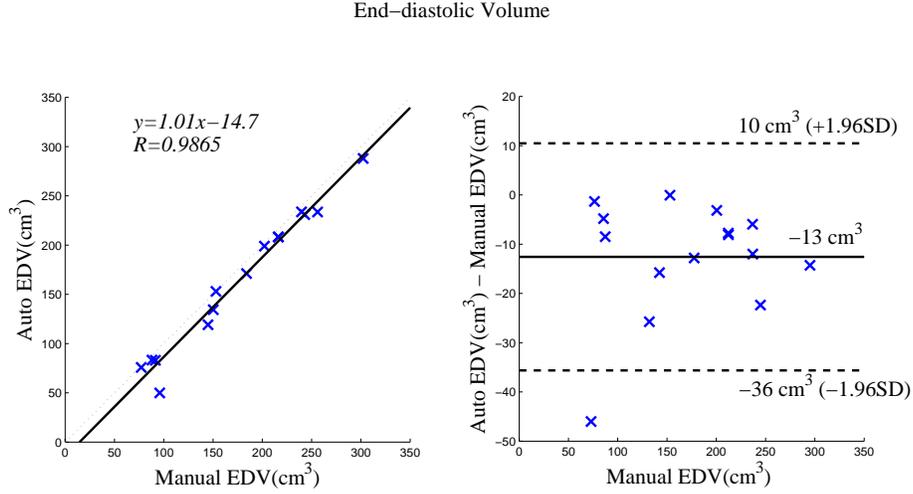},width=\linewidth}}
\caption{Correlation graph (left) and Bland-Altman(right) for end-diastolic volume (EDV).}
\label{fig:bland-edv}
\end{figure}
\begin{figure}[h]
\centerline{\epsfig{figure={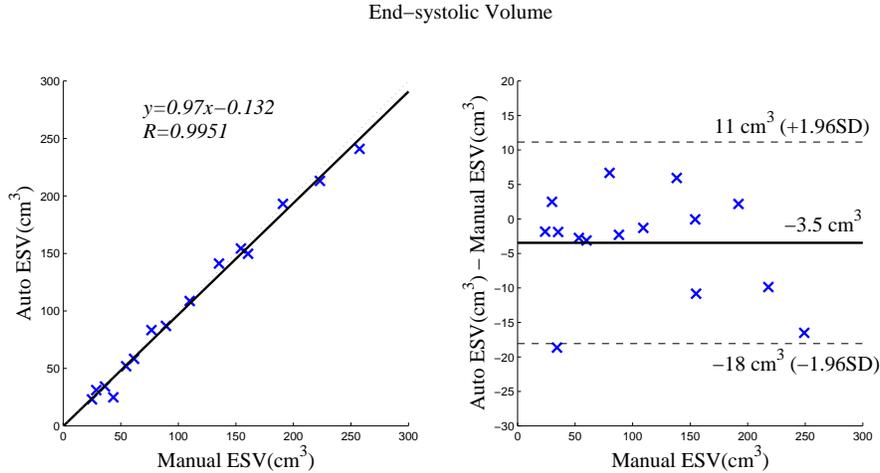},width=\linewidth}}
\caption{Correlation graph (left) and Bland-Altman(right) for end-systole volume (ESV).}
\label{fig:bland-esv}
\end{figure}

\begin{figure}[h]
\psfrag {Ejection Fraction} [c] [] [1.0] {}
\psfrag {Mean} [c] [] [1.0] {}
\centerline{\epsfig{figure={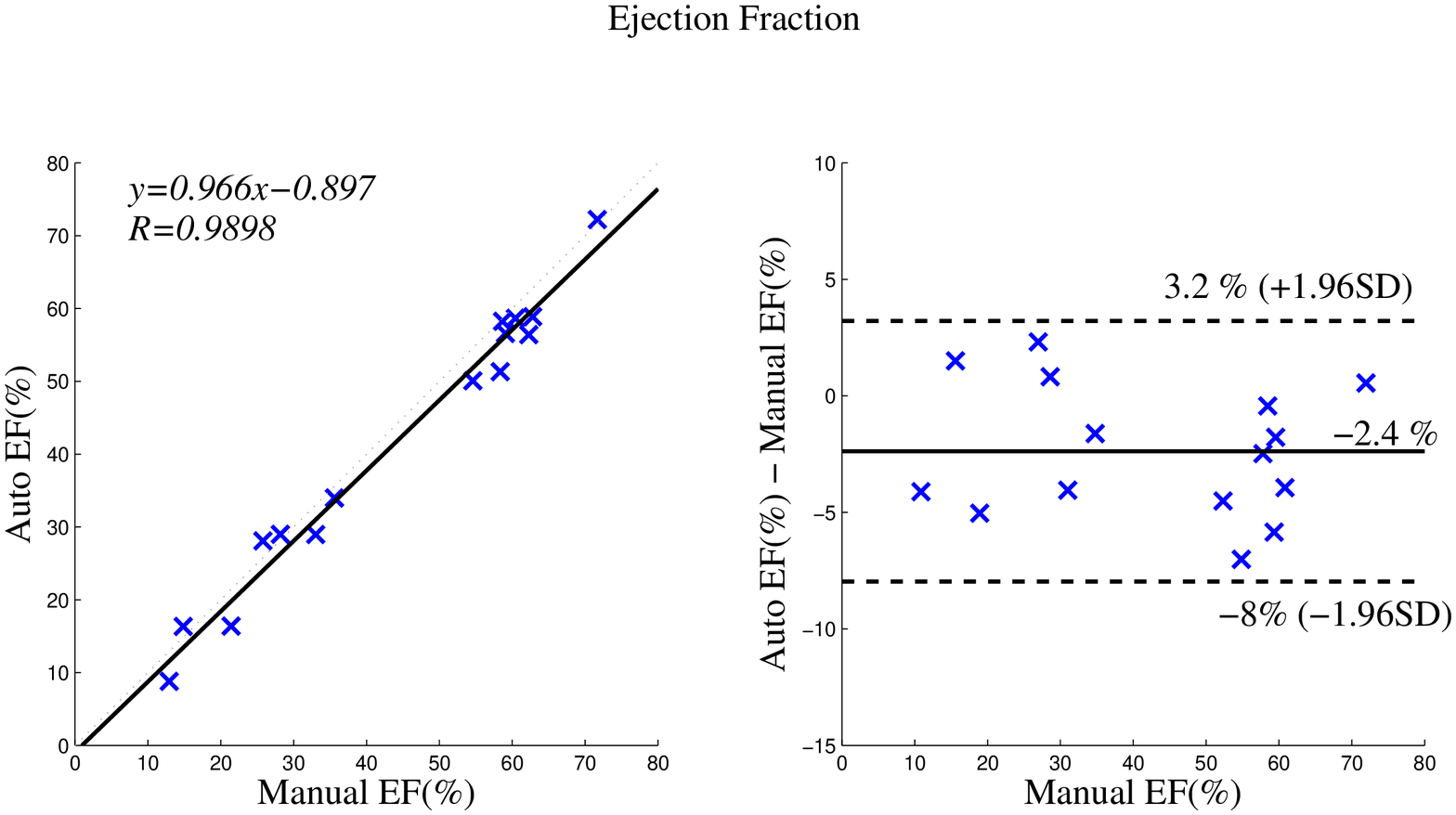},width=\linewidth}}
\caption{Correlation graph (left) and Bland-Altman(right) for the ejection fraction (EF).}
\label{fig:bland-ef}
\end{figure}

Moreover, Figs.~\ref{fig:bland-edv}-\ref{fig:bland-ef} illustrate the correlation graphs (left) between the automatic and manual results and the Bland-Altman graphs (right) of the differences, using the validation dataset, for EDV, ESV and EF, respectively. A correlation with the ground truth contours of 0.99, 0.99, 0.99 for EDV, ESV and EF was measured. The level of agreement between the automatic and manual results was represented by the interval of the percentage difference between mean$\pm 1.96$SD. The mean and confidence interval of the difference between the automatic and manual EDV results were -13 cm$^3$   and (-36cm$^3$ to 10cm$^3$), respectively. The CV and RPC were 6.9\% and 13\%, respectively. The mean and confidence interval of difference between the automatic and manual ESV results were -3.5 cm$^3$, (-18cm$^3$ to 11cm$^3$) and CV=6.9\%, RPC=14\%.  Also, the mean and confidence interval of the difference between the automatic and manual EF results were -2.4\%, (-8\% to 3.2\%), CV=6.7\%, RPC=13\%.

Approximated elapsed times of the training process were as follows. Training autoencoder to obtain filters: 63.3 seconds, training convolutional network: 3.4 hours, training stacked-AE: 34.25 minutes. Once trained, the elapsed times of segmenting the LV in a typical MR image were as follows: ROI detection (convolution, pooling, and logistic regression): 0.25 seconds, shape inferring (stacked-AE): 0.002 seconds, segmentation (deformable model): 0.2 seconds.

\section{Discussion}
\label{sec:diss}
In this study, we developed and validated an automated segmentation method for the LV based on deep learning algorithms. We broke down the problem into localization, shape inferring and segmentation tasks. Convolutional networks were chosen for localization and extracting an ROI because they are invariant to spacial translation and changes in scale and pixels' intensity \citep{cnn-lecun,cnnlecunn2014}. We also chose a stacked AE for shape inferring because of its simplicity in training and implementation yet showing to be powerful in different vision tasks. \citep{vincent2010stacked}. Ideally, a pure deep learning was desired. However, this was not possible due to several challenges including the limited amount of training data. Thus, we integrated deep learning with deformable models to bring more accuracy to the method. 

\addt{
As seen in the left side of Fig.~\ref{fig:each_step}, the outcome of the deformable model without shape constraint (blue) leaked to surrounding tissues due to low contrast at the borders. Clearly this is not acceptable. On the other hand, the deep learning network (shape inference) provided a close contour (red) to the ground truth (green) with no leakage. This is due to the fact that the network has been trained using the ground truth data to look for the overall shape of the LV and not the intensity difference at the border. Finally, the integrated deep learning and deformable models brought the contour (yellow) closer to the ground truth. Similar behavior can be seen in the right side of Fig.~\ref{fig:each_step} when contours tend to shrink due to presence of papillary muscles in the LV.
}

From Figs.~\ref{fig:segs2d3d}-\ref{fig:seg_multiplecases}, it can be seen that the LV was accurately segmented from the base to the apex. The alignment process resulted in a smooth 3D reconstruction of the LV in Fig.~\ref{fig:segs2d3d}. The first image on the left corner of Fig.~\ref{fig:segs2d3d} shows a slight leakage from the ground truth. This situation is one of the challenging cases that, due to the fuzziness of the LV border, contours tend to leak to surrounding tissues in pure deformable models. Luckily, by integrating the inferred shape into the deformable models, this leakage was significantly prevented in this image and also all similar cases in the dataset such as the first and second images in the fifth row of Fig.~\ref{fig:seg_multiplecases}. In other challenging cases, such as in Fig.~\ref{fig:challenging}, that pure deformable models tend to shrink inward due to the presence of papillary muscles, or leak outward due to low resolution and small contrast of images at the apex, our method overcame these shortcomings.

Computed metrics in Table~\ref{tab:comp1} showed that the inferred shape provided good initial contours with an overall accuracy of 90\% (in terms of DM). Also, the integrated deformable model provided final contours with great agreement with the ground truth with an overall DM of 94\% and improvements in other metrics. Table~\ref{tab:comp2} revealed that our method outperformed the state-of-the-art methods and significant improvements were achieved in all metrics. Specifically, the DM and conformity were improved by 4\%  and 0.08 compared to the best DM and conformity reported by \citep{queiros-2014}.  

The correlation analysis in Figs.~\ref{fig:bland-edv}-\ref{fig:bland-ef} depicted a high correlation for the three clinical cardiac indices. The high correlation between the automatic and manual references shows the accuracy and clinical applicability of the proposed framework for automatic evaluation of the LV function. Also, the Bland-Altman analysis in the figures revealed a better level of agreement compared with that of \citep{queiros-2014}. On the other hand, the level of agreement of frameworks of \citep{CorderoGrande2011283,Eslami2013236} are slightly better than that of our method, which can be related to the semi-automated property of these methods compared with our fully automatic approach.

The measured elapsed time revealed that the method can be trained within a reasonable time, which can be performed offline. The longest time was needed for the convolutional network, which required convolution of the filters with images. Nevertheless, these times can be even shortened by developing the algorithms into GPU-accelerated computing platforms instead of our current CPU-based platform. In testing, the average time to perform the LV segmentation in a typical image was found less than 0.5 seconds, of which mostly taken by the convolution network and the integrated deformable model. Yet, the integrated deformable model converges faster than pure deformable models because of the initialization and integration with the inferred shape. Some of the published works provide numbers for the computational time. However, since each method has been developed on a different computing platform, these values may not be reliable for comparison unless all the methods are developed on the same platform.

\addt{
It is noted that, while 3D methods are becoming the state-of-the-art in many medical image analysis applications, we performed 2-dimensional (2D) processing in the present study. This choice was due to two known challenges in cardiac MRI that prevents one from direct 3-dimensional (3D) analysis. First, the gap between slices (vertical dimension) in most routine acquisitions is relatively large (around 7-8 mm) and the pixel intensities between the slices cannot be reliably estimated \citep{review-petit,review-tavakoli,Petitjean2015187,queiros-2014}. Second, due to motion artifacts in cardiac MRI, misalignment between slices is common \citep{misalign2006,misalign2008,misalign2004,misalign2012,misalign2014,misalign2015}. This means that the cavity center is not at the same position in different slices. Some of existing tools perform an initial 2D segmentation in the middle slice and later apply an alignment process and then convert from the Cartesian coordinate to the polar coordinate to be able to perform 3D processing \citep{queiros-2014}. Alternatively, atlas-based techniques build a reference 3D model from some training data and then register the new image to the reference model, which limits the accuracy of segmentation to the reference model. Accordingly, different approaches can be adapted for our method if 3D processing is sought. For instance, instead of feeding 2D images in the current method, 3D data can be fed to the deep learning networks and trained for a 3D reconstruction. This would require networks with higher number of nodes and layers. Considering these challenges and additional complexity burden, the possibility of performing 3D computation can be investigated in future. Nevertheless, cardiac chamber segmentation in clinical practice is mainly used to compute clinical indices such as the volume or ejection fraction. Our proposed method is able to provide these indices with high accuracy while performing 2D processing. }


Finally, one of the difficulties in developing deep learning and machine learning approaches for cardiac MRI segmentation is the lack of adequate data for training and validation. Particularly for deep learning networks, access to more data helps to reach a better generalization and reduce the overfitting problem. For this work, we used a portion of the MICCAI dataset \citep{miccai2009} and artificially enlarged the dataset for training. However, in this case the training data are highly correlated, which would limit the performance. Also, currently, there are no analytic approaches to design hyper-parameters (such as number of layers and units, filter size, etc.) in deep learning networks and they are mainly obtained empirically, as we performed in our study. 

\section{Conclusion}
\label{sec:con}
In summary, a novel method for fully automatic segmentation of the LV from cardiac MRI datasets was presented. The method employed deep learning algorithms for automatic detection and inferring the shape of the LV. The shape was incorporated into deformable models and brought more robustness and accuracy, particularly for challenging basal and apical slices. 
The proposed approach was shown to be accurate and robust compared to the other state-of-the-art methods.
Excellent agreement and a high correlation with reference contours were obtained. In contrast with other automated approaches, our method is based on learning several levels of representations, corresponding to a hierarchy of features and does not assume any model or assumption about the image or heart. The feasibility and performance of this segmentation method was successfully demonstrated through computing validation metrics with respect to the gold standard on the MICCAI 2009 database \citep{miccai2009}. 
Testing our method on a larger set of clinical data is subject of future research. 

\section*{Acknowledgments}
This work is partially supported by a grant from American Heart Association (14GRNT18800013).

\section*{References}
\bibliographystyle{model2-names}
\bibliography{ref/references}

\end{document}